\documentclass{article}

\usepackage[preprint]{neurips_2023}

\usepackage{natbib}
\setcitestyle{numbers,square}

\usepackage[utf8]{inputenc} 
\usepackage[T1]{fontenc}    
\usepackage{hyperref}       
\usepackage{url}            
\usepackage{booktabs}       
\usepackage{amsfonts}       
\usepackage{nicefrac}       
\usepackage{microtype}      
\usepackage{xcolor}         

\usepackage{times}
\usepackage{epsfig}
\usepackage{graphicx}
\usepackage{amsmath}
\usepackage{amssymb}

\usepackage{amsmath}
\usepackage{amssymb}
\usepackage{mathtools}
\usepackage{amsthm}

\usepackage{multirow}

\usepackage{algorithm}
\usepackage[]{algpseudocode}

\title{ High Quality Diffusion Distillation on a Single GPU with Relative and Absolute Position Matching }

\author{%
  Guoqiang Zhang  \\
  University of Exeter \\
  \texttt{g.z.zhang@exeter.ac.uk} \\
     \And  Kenta Niwa \\
    NTT Communication Science Laboratories \\
   \texttt{kenta.niwa@ntt.com}  \\
   \And  J.P. Lewis
    \\
   Victoria University of Wellington \\
\texttt{noisebrain@gmail.com} \\
   \And  Cedric Mesnage
    \\
   University of Exeter \\
\texttt{c.s.mesnage@exeter.ac.uk} \\
   \And
   W. Bastiaan Kleijn \\ 
Victoria University of Wellington \\ \texttt{bastiaan.kleijn@vuw.ac.nz} \\
}

\begin{document}

\maketitle

\begin{abstract}

We introduce relative and absolute position matching (RAPM), a diffusion distillation method resulting in high quality generation that can be trained efficiently on a single GPU.
Recent diffusion distillation research has achieved excellent results for high-resolution text-to-image generation with methods such as phased consistency models (PCM) and improved distribution matching distillation (DMD2). However, these methods generally require many GPUs (e.g.~8-64) and significant batchsizes (e.g.~128-2048) during training, resulting in memory and compute requirements that are beyond the resources of some researchers.
RAPM  provides effective single-GPU diffusion distillation training with a batchsize of 1. The new method attempts to mimic the sampling trajectories of the teacher model by matching the relative and absolute positions. The design of relative positions is inspired by PCM. Two discriminators are introduced accordingly in RAPM, one for matching relative positions and the other for absolute positions. Experimental results on StableDiffusion (SD) V1.5 and SDXL indicate that RAPM with 4 timesteps produces comparable FID scores as the best method with 1 timestep under very limited computational resources. 

\end{abstract}

\section{Introduction}
\label{sec:intro}

As one type of generative models \cite{Goodfellow14GAN, Arjovsky17WGAN, Gulrajani17WGANGP, Sauer22StyleGAN, Bishop06}, diffusion probabilistic models (DPMs) have made remarkable progress in the past few years. Following the  pioneering work of \cite{Dickstein15DPM}, various training and/or sampling strategies have been proposed to improve the performance of DPMs, which include, for example, denoising diffusion probabilistic models (DDPMs) \cite{Ho20DDPM}, denoising diffusion implicit models (DDIMs) \cite{Song21DDIM}, latent diffusion models (LDMs)\cite{Rombach22LDM}, score matching with Langevin dynamics (SMLD) \cite{Song19, Song21DPM, Song21SDE_gen}, analytic-DPMs \cite{Bao22DPM, Bao22DPM_cov},  optimized denoising schedules \cite{Kingma21DDPM, Chen20WaveGrad, Lam22BDDM},  classifier-free guided diffusion \cite{Ho22ClassiferFreeGuide}, and guided diffusion strategies \cite{Nichol22GLIDE, Kim22GuidedDiffusion}.

In general, a standard diffusion model needs to perform a considerable number of neural functional evaluations (NFEs) to generate high quality images, which is computationally expensive and time-consuming. One approach to reduce NFEs is to design and exploit high-order ordinary differential equation (ODE) solvers in diffusion sampling.  It is worth noting that DDIM can be interpreted as a first-order ordinary differential equation (ODE) solver. As an extension of DDIM, various high-order ODE solvers have been proposed, such as EDM \cite{Karras22EDM}, DEIS \cite{Zhang22DEIS}, PLMS and PNDM \cite{Liu22PNDM}, DPM-Solvers++ \cite{Lu22DPM_SolverPlus}, IIA-EDM and IIA-DDIM \cite{GuoqiangIIA23}, and BDIA-DDIM \cite{Zhang24BDIA}. Generally speaking, high-order ODE solvers do not perform well for very small NFEs (e.g., from 1 to 4 NFEs) due to their limited capabilities. 

\begin{table}[t]
\caption{Computational resources of representative diffusion distillation methods  for stable diffusion (SD) V1.5. We emphasize that no gradient accumulation is required for RAPM. }
\centering
\label{tab:GPU_bachsize}
\begin{tabular}{|c|c|c|c|c|}
\hline
    &  {\footnotesize GPUs} 
    \hspace{-2mm} & \hspace{-2mm} {\footnotesize batchsize} 
    \hspace{-2mm} & \hspace{-2mm} 
    {\footnotesize $\begin{array}{c}\textrm{training time} \\ \textrm{(hours)}\end{array}$} 
    \hspace{-2mm} & \hspace{-2mm} {\footnotesize LoRA} \\
    \hline
     {\footnotesize SFD \cite{Zhou24SFD} } \hspace{-2mm} 
     \hspace{-2mm} & \hspace{-2mm} \hspace{-2mm} 
     {\footnotesize 4 A100} 
     \hspace{-2mm} & \hspace{-2mm} 
     {\footnotesize  128} 
     \hspace{-2mm} & \hspace{-2mm} {\footnotesize -} \hspace{-2mm} & \hspace{-2mm} {\footnotesize No } \\    
    \hline
     {\footnotesize DMD2 \cite{Yin24DMD}} \hspace{-2mm} & \hspace{-2mm} {\footnotesize 64 A100} \hspace{-2mm} & \hspace{-2mm} {\footnotesize 2048} \hspace{-2mm} & \hspace{-2mm} {\footnotesize 26}   \hspace{-2mm} & \hspace{-2mm} {\footnotesize No } \\
     \hline
    {\footnotesize PCM \cite{Wang24PCM} }  \hspace{-2mm} & \hspace{-2mm} {\footnotesize 8 A800}  
    \hspace{-2mm} & \hspace{-2mm} {\footnotesize 160}  
    \hspace{-2mm} & \hspace{-2mm}
    {\footnotesize -}  
    \hspace{-2mm} & \hspace{-2mm} {\footnotesize Yes } \\
     \hline
     {\footnotesize RAPM (\textbf{our})} \hspace{-2mm} & \hspace{-2mm} {\footnotesize 1 A6000}  
     \hspace{-2mm} & \hspace{-2mm}  {\footnotesize 1} 
     \hspace{-2mm} & \hspace{-2mm} 
     {\footnotesize 8.3}  
     \hspace{-2mm} & \hspace{-2mm} {\footnotesize Yes }  \\  
     \hline
\end{tabular}
\end{table}

\begin{figure*}[t!]
\centering
\includegraphics[width=140mm]{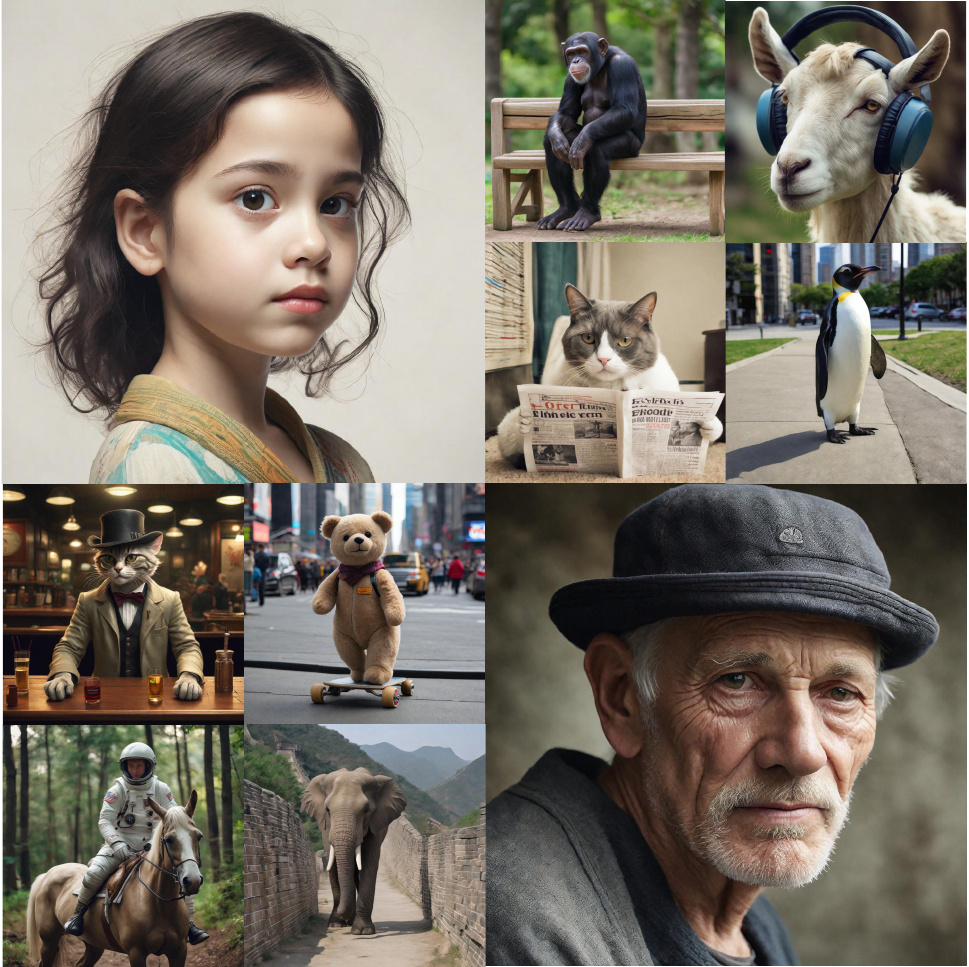} 
\caption{1024$\times$1024 samples produced by our 4-step generator distilled from SDXL. See Table~\ref{tab:prompts_RAPM_images} for the text prompts. } 
\label{fig:RAPM_images}
\vspace*{-0.0cm}
\end{figure*}

To further reduce NFEs with little performance degradation, the diffusion distillation approach has attracted increasing attention in the literature via the training of a student diffusion model from a teacher model. For example, the work of \cite{Salimans22DDIMDis} proposes progressive distillation to train a sequence of student models by using the DDIM sampling method, where the final student model aims to produce satisfactory image quality with a very small NFE.  The authors of \cite{Zhou24SiD} propose score identity distillation (SiD) that uses synthetic data of the student model in the training process. However, SiD is only evaluated for conventional image generation. It is not clear how good the performance of SID is for text-to-image generation.  \cite{Zhou24SiD}.  Other representative distilation methods include, for example,  consistency models \cite{Song23CM, Song23CMb}, and consistency trajectory models \cite{Kim23}.

Recent research activities of diffusion distillation focus on high-resolution text-to-image generation.   Typical distillation methods in the literature include, for example, simple and fast distillation (SFD) \cite{Zhou24SFD}, phased consistency models (PCM) \cite{Wang24PCM}, distribution distillation matching (DMD) \cite{yin2024onestep}, and improved DMD (referred to as DMD2) \cite{Yin24DMD}. However, all of the above four methods require quite a number of high-end GPUs to accommodate reasonable batch sizes of training data per iteration (see Table~\ref{tab:GPU_bachsize} for details). As shown in the table, PCM utilizes the LoRA adapter \cite{Hu21LoRA} to reduce the number of trainable parameters in the student diffusion model, which is different from other distillation methods.

Next we briefly explain the principle of PCM. For each considered sampling time-slot, PCM computes two separate integration approximations via DDIM, one with only the learnable student model and the other with both the teacher and the frozen student model. Real images contaminated with Gaussian noise are fed into the diffusion models when computing the integration approximations. PCM then attempts to minimize the distance between the two separate integration approximations in terms of both a Huber loss and a GAN-loss. As shown in Table~\ref{tab:GPU_bachsize}, PCM needs 8 A800 GPUs to accommodate the training data with batchsize of 160 for SD V1.5.   

In this paper, we propose a new distillation method, named \emph{relative and absolute position matching} (RAPM), for high-quality diffusion distillation on a single GPU. To allow for the extreme training setup of batchsize-1 per iteration,  we first obtain the diffusion trajectory of the teacher model for an initial Gaussian noise over a set of fine-grained discrete timesteps. The same Gaussian noise is fed into the student model over a set of coarse timesteps (e.g., 2 or 4 timesteps). The student model is then trained by matching the student diffusion positions with  the relative and absolute positions obtained by utilizing both the teacher and the frozen student model. As will be explain in Section~\ref{sec:RAPM}, computation of the relative diffusion position is inspired by PCM. Two discriminators are introduced in RAPM, one for matching the relative diffusion positions and the other one for the absolute diffusion positions. The losses from the two discriminators are complementary to each other for the student model to closely mimic the diffusion trajectories of the teacher model. Similar to PCM, RAPM also utilizes the LoRA adapters in both the student model and the discriminators  to reduce the number of trainable parameters. 

We make two contributions towards high-quality diffusion distillation over a single GPU. Firstly, proper Huber losses and GAN-losses are carefully designed for matching the diffusion positions mentioned above. We note that naive construction of the losses may lead to degraded performance.  Secondly, extensive experiments show that RAPM with 4 timesteps yields comparable FID performance to the best method with 1 timestep using very limited computational resources when distilling SD V1.5 and SDXL. 

\vspace{-0mm}
\section{Preliminaries}
    \label{sec:pre}
\vspace{-0mm}
\textbf{Forward and reverse diffusion processes:}
Suppose a data sample $\boldsymbol{x}\in \mathbb{R}^d$ are drawn from a data distribution $p_{data}(\boldsymbol{x})$ with a finite variance. A forward diffusion process progressively inserts Gaussian noise to the data samples $\boldsymbol{x}$ to obtain $\boldsymbol{z}_t$ as $t$ increases from 0 until $T$. The conditional distribution  of $\boldsymbol{z}_t$ given $\boldsymbol{x}$ can be expressed as
\begin{align}
q_{t|0}(\boldsymbol{z}_t|\boldsymbol{x}) = \mathcal{N}(\boldsymbol{z}_t|\alpha_t\boldsymbol{x}, \sigma_t^2\boldsymbol{I})\quad \boldsymbol{z}_t = \alpha_t\boldsymbol{x}+\sigma_t \boldsymbol{\epsilon},
\label{equ:forwardGaussian}
\end{align}
where $\alpha_t$ and $\sigma_t$ are differentiable functions of $t$ with finite derivatives. Suppose that the three functions $\alpha_t$, $\sigma_t$, and $\lambda_t=\frac{\sigma_t}{\alpha_t}$, are non-increasing, non-decreasing,  and strictly
increasing (see \cite{Lu22DPM_Solver}) as $t$ increases, respectively. We let $q(\boldsymbol{z}_t; \alpha_t,\sigma_t)$ denote the marginal distribution of $\boldsymbol{z}_t$. In principle, the samples drawn from the distribution $q(\boldsymbol{z}_T;\alpha_T,\sigma_T)$ should be indistinguishable from pure Gaussian noise if $\sigma_T \gg \alpha_T$.

The reverse diffusion process firstly draws a sample $\boldsymbol{z}_T$ from $\mathcal{N}(\boldsymbol{0}, {\sigma}_{T}^2\boldsymbol{I})$, and then progressively denoises it to obtain a sequence of diffusion states (or diffusion positions)  $\{\boldsymbol{z}_{t_i}\sim p(\boldsymbol{z};\alpha_{t_i},\sigma_{t_i})\}_{i=0}^N$,
where the notation $p(\cdot)$ is used to indicate that the reverse sample distribution might be different from $q(\cdot)$ because of  approximation errors.  The research objective is to compute a final sample $\boldsymbol{z}_{t_0}$ which is roughly distributed according to $p_{data}(\boldsymbol{x})$. 

\noindent\textbf{ODE formulation:} The work  \cite{Song21SDE_gen} presents a so-called \emph{probability flow} ODE which shares the same marginal distributions as $\boldsymbol{z}_t$ in (\ref{equ:forwardGaussian}).  Given the formulation (\ref{equ:forwardGaussian}) for a forward diffusion process, its reverse ODE takes the form of \cite{Lu22DPM_Solver, GuoqiangIIA23}
\begin{align}
d\boldsymbol{z} \hspace{-1mm}=&\hspace{-1mm} \overbrace{\left[\hspace{-1mm}\frac{d\log \hspace{-0.6mm}\alpha_t}{dt}\boldsymbol{z}_t \hspace{-1mm}-\hspace{-1mm}\frac{1}{2}\hspace{-1mm}\left(\hspace{-0.6mm}\frac{d\sigma_t^2}{dt} \hspace{-1mm}-\hspace{-1mm} 2\frac{d\log\alpha_t}{dt}\sigma_t^2\hspace{-0.6mm}\right)\hspace{-1mm}\nabla_{\boldsymbol{z}}\hspace{-1mm}\log q(\boldsymbol{z}_t; \alpha_t,\sigma_t)\hspace{-0.6mm}\right]}^{\boldsymbol{d}(\boldsymbol{z}_t,t)}\hspace{-1mm}dt,
\label{equ:ODE_gen}
\end{align}
where 
$\nabla_{\boldsymbol{z}}\log q(\boldsymbol{z};\alpha_t,\sigma_t)$ in (\ref{equ:ODE_gen}) is the score function  \cite{Hyvarinen05ScoreMatching} pointing towards higher density of data samples at the given noise level $(\alpha_t,\sigma_t)$. It is noted that the score function does not depend on the intractable normalization constant of the underlying density function $q(\boldsymbol{z};\alpha_t,\sigma_t)$, which make the denoising process much easier. 

As $t$ decreases towards 0, the ODE formulation (\ref{equ:ODE_gen}) continuously reduces the noise level of the data samples in the reverse process. In the ideal case where no approximations are introduced in the ODE (\ref{equ:ODE_gen}) at all, the reverse sample distribution $p(\boldsymbol{z};\alpha_t,\sigma_t)$ approaches the original data distribution $p_{data}(\boldsymbol{x})$ as $t$ goes from $T$ to 0. In other words, the denoising process of a diffusion model boils down to solving the ODE form (\ref{equ:ODE_gen}), where only the initial sample at time $T$ introduces randomness. As mentioned in the introduction, significant progress has been made by exploiting or designing different ODE solvers in diffusion-based sampling processes.     

\noindent\textbf{Denoising score matching:} One needs to specify a particular form of the score function $\nabla_{\boldsymbol{z}}\log q(\boldsymbol{z};\alpha_t,\sigma_t)$ before utilizing (\ref{equ:ODE_gen}) for sampling.  The most popular scheme is to train a noise estimator $ \hat{\boldsymbol{\epsilon}}_{\boldsymbol{\theta}}$ by minimizing the expected $L_2$ error for samples drawn from $q_{data}$ (see \cite{Ho20DDPM, Song21SDE_gen, Song21DDIM}):
\begin{align}
\mathbb{E}_{\boldsymbol{x}\sim p_{data}}\mathbb{E}_{\boldsymbol{\epsilon}\sim \mathcal{N}(\boldsymbol{0}, \sigma_t^2\boldsymbol{I})}\|\hat{\boldsymbol{\epsilon}}_{\boldsymbol{\theta}}(\alpha_t \boldsymbol{x}+\sigma_t\boldsymbol{\epsilon},t)\hspace{-0.3mm}-\hspace{-0.3mm}\boldsymbol{\epsilon}\|_2^2,\label{equ:epsilon_training}
\end{align}
where $(\alpha_t, \sigma_t)$ are as specified in the forward diffusion process (\ref{equ:forwardGaussian}).  A typical approach in diffusion models is to employ a neural network based on the U-Net architecture \cite{Ronneberger15Unet} to model the noise estimator $\hat{\boldsymbol{\epsilon}}_{\boldsymbol{\theta}}$.
Finally, by utilizing (\ref{equ:epsilon_training}), the score function (\ref{equ:ODE_gen}) can be represented in terms of 
$\hat{\boldsymbol{\epsilon}}_{\boldsymbol{\theta}}(\boldsymbol{z}_t; t)$ as (see also (229) of \cite{Karras22EDM})
\begin{align}
\nabla_{\boldsymbol{z}}\log q(\boldsymbol{z}_t;\alpha_t,\sigma_t) \hspace{-0.0mm}=\hspace{-0.0mm} {-(\boldsymbol{z}_t\hspace{-0.0mm}-\hspace{-0.0mm}\alpha_t \boldsymbol{x})}/{\sigma_t^2}  \hspace{-0.0mm}=\hspace{-0.0mm} -\hat{\boldsymbol{\epsilon}}_{\boldsymbol{\theta}}(\boldsymbol{z}_t; t)/\sigma_t. \label{equ:score_gaussian}
\end{align}

\section{Relative and Absolute Position Matching (RAPM)}
\label{sec:RAPM}

In this section, we first briefly review the update procedure of PCM \cite{Wang24PCM}, which provides inspiration for our method. We then present our new distillation method RAPM.  In particular, we explain in detail how to match  the relative and absolute diffusion positions when updating the student model. The impact of the two matching schemes on image quality will also be studied.  Lastly, we inspect the similarities and differences between RAPM and SFD \cite{Zhou24SFD}.

\subsection{Revisiting PCM}
In general, PCM attempts to learn a student diffusion model $\boldsymbol{\varphi}$ from a teacher model  $\boldsymbol{\theta}$ such that after training, the student model produces high-quality images over a set of coarse timesteps $t_N>t_{N-1}>\cdots> t_0$, where typical values of $N$ are from 1 to 4. While the consistency model \cite{Song23CM} enforces consistency between the teacher and student model from the perspective of the global time-scale $[0,T]$, PCM enforces consistency within each sampling time-slot $(t_{n+1}, t_{n}]$, $n=0,\ldots, N{-}1$.

 \begin{algorithm}[tb]
   \caption{ PCM \cite{Wang24PCM}}
   \label{alg:PCM}
\begin{algorithmic}[1]
   \State \textbf{Input}: $\boldsymbol{\theta}$: teacher model, $\{t_n\}_{n=N}^{0}$, $\Delta$
\State $\;\;\;\;\;\;\;\;\;$ $\boldsymbol{\varphi}$: LoRA parameters in student model  
\State $\;\;\;\;\;\;\;\;\;$ $D_{\boldsymbol{\phi}_1}$: discriminator 1  
   \State \textbf{for} $k=0,1,\ldots,K$ 
   \State $\;\;$ Sample $\boldsymbol{x}$ from the dataset
   \State $\;\;$ Sample $t\sim \mathcal{U}(0,T)$
    \State $\;\;$ Sample $\boldsymbol{z}_t\sim \mathcal{N}(\alpha_t\boldsymbol{x}; \sigma_{t}^2\boldsymbol{I})$
   \State $\;\;$ Find $t_n$ such that $t_{n+1}> t-\Delta > t_n$ 
    \State $\;\;$ $\hat{\boldsymbol{z}}_{t}^{\Delta}\leftarrow \textrm{DDIM}(\boldsymbol{z}_{t}, t, t-\Delta, 1; \boldsymbol{\theta})$ 
    \State $\;\;$ $\hat{\boldsymbol{z}}_{n}^{\boldsymbol{\varphi}}\leftarrow \textrm{DDIM}(\hat{\boldsymbol{z}}_{t}^{\Delta}, t-\Delta, t_{n}, 1; \boldsymbol{\varphi})$
    \State $\;\;$ $\hat{\boldsymbol{z}}_{n}\leftarrow \textrm{detach}(\hat{\boldsymbol{z}}_{n}^{\boldsymbol{\varphi}})$
\State $\;\;$ $\boldsymbol{z}_n^{\boldsymbol{\varphi}}\leftarrow \textrm{DDIM}(\boldsymbol{z}_{t}, t, t_n, 1; \boldsymbol{\varphi})$    
    \State $\;\;$ $\textrm{loss }\hspace{-1mm}= H(\boldsymbol{z}_n^{\boldsymbol{\varphi}}, \hat{\boldsymbol{z}}_{n})+D_{\boldsymbol{\phi}_1}(\boldsymbol{z}_n^{\boldsymbol{\varphi}}, \hat{\boldsymbol{z}}_{n}|t_n) $
    \State $\;\;$ \textbf{if} $\textrm{mod}(k,2)=0$
    \State $\;\;\;\;$ $\boldsymbol{\phi}_1\leftarrow \boldsymbol{\phi}_1 - \eta\nabla_{\boldsymbol{\phi}_1 }  \textrm{loss}$ $\;\;\;\;$ \textcolor{blue}{[update discriminator]}
    \State \;\;\textbf{else}
    \State $\;\;\;\;$ $\boldsymbol{\varphi}\leftarrow \boldsymbol{\varphi} - \eta\nabla_{\boldsymbol{\varphi} }  \textrm{loss}$ $\;\;\;\;\;\;\;\;$  \textcolor{blue}{[update student model]}
  \State  \textbf{end for}
\end{algorithmic}
\vspace{-0.5mm}
\end{algorithm}

Next we briefly explain how the consistency is enforced within a particular sampling time-slot $(t_{n+1}, t_{n}]$. Suppose $t$ is sampled randomly from $\mathcal{U}[0,T]$ and satisfies $t_{n+1}>t{-}\Delta\geq t_n $, where $\Delta>0$ is a small value. Next, a real image $\boldsymbol{x}$ is selected randomly from the dataset. With $\boldsymbol{x}$, one can then construct a noisy image $\boldsymbol{z}_t\sim \mathcal{N}(\alpha_t\boldsymbol{x}; \sigma_{t}^2\boldsymbol{I})$ by following (\ref{equ:forwardGaussian}). Upon introducing $\boldsymbol{z}_t$,  one can simply make use of the student model $\boldsymbol{\varphi}$ and apply 1-step DDIM over the time-slot $[t, t_n]$ to compute $\boldsymbol{z}_n^{\boldsymbol{\varphi}}\equiv\boldsymbol{z}_{t_n}^{\boldsymbol{\varphi}}$, given by:   
\begin{align}
\boldsymbol{z}_n^{\boldsymbol{\varphi}} = \textrm{DDIM}(\boldsymbol{z}_{t}, t, t_n, 1; \boldsymbol{\varphi}),
\label{equ:PCM_S}
\end{align}
where $\boldsymbol{\varphi}$ is to be learned. 

To be able to provide guidance for updating $\boldsymbol{\varphi}$, PCM then computes a more accurate diffusion position $\hat{\boldsymbol{z}}_{n}$ by performing two 1-step DDIM evaluations by using the teacher and the student model sequentially over the time-slots $[t,t{-}\Delta]$ and $[t{-}\Delta, t_n]$, represented by
\begin{align}
\hat{\boldsymbol{z}}_{t}^{\Delta} &= \textrm{DDIM}(\boldsymbol{z}_{t}, t, t-\Delta, 1; \boldsymbol{\theta}) \label{equ:PCM_T1} \\
\hat{\boldsymbol{z}}_{n}^{\boldsymbol{\varphi}} &= \textrm{DDIM}(\hat{\boldsymbol{z}}_{t}^{\Delta}, t-\Delta, t_{n}, 1; \boldsymbol{\varphi})  \label{equ:PCM_T2} \\
 \hat{\boldsymbol{z}}_{n} &= \textrm{detach}(\hat{\boldsymbol{z}}_{n}^{\boldsymbol{\varphi}}),  \label{equ:PCM_T3}
\end{align}
where $\hat{\boldsymbol{z}}_{t}^{\Delta}$ is the diffusion position at timestep $t-\Delta$ obtained by using the teacher model. The detach operation in (\ref{equ:PCM_T3}) makes $ \hat{\boldsymbol{z}}_{n}$ independent of $\boldsymbol{\varphi}$, and is equivalent to applying the frozen student model when computing $\hat{\boldsymbol{z}}_{n}$. In principle, the diffusion position $\hat{\boldsymbol{z}}_{n}$ should be more accurate than $\boldsymbol{z}_n^{\boldsymbol{\varphi}}$ in (\ref{equ:PCM_S}) since  $\hat{\boldsymbol{z}}_{n}$ is obtained by applying two 1-step DDIM evaluations. 

In \cite{Wang24PCM}, the difference between $\hat{\boldsymbol{z}}_{n}$ and  $\boldsymbol{z}_n^{\boldsymbol{\varphi}}$ is measured in terms of a Huber-loss and a GAN-loss, given by 
\begin{align}\textrm{loss}(\boldsymbol{\phi}_1,\boldsymbol{\varphi} )\hspace{-1mm}= H(\boldsymbol{z}_n^{\boldsymbol{\varphi}}, \hat{\boldsymbol{z}}_{n})+D_{\boldsymbol{\phi}_1}(\boldsymbol{z}_n^{\boldsymbol{\varphi}}, \hat{\boldsymbol{z}}_{n}|t_n),
\label{equ:PCM_loss}
\end{align}
where $H(\cdot,\cdot)$ refers the Huber-loss and $D_{\boldsymbol{\phi}_1}$ represents a discriminator. As training continues, the discriminator and the student model are updated alternatingly by minimizing the loss function in (\ref{equ:PCM_loss}). See Alg.~\ref{alg:PCM} for the update procedure.    

We note that in \cite{Wang24PCM}, both the PCM student model and the discriminator $D_{\boldsymbol{\phi}_1}$ are designed by introducing LoRA adapters into the teacher model. Therefore, the number of trainable parameters of PCM is significantly smaller than that of competitive distillation methods.

\begin{algorithm}[tb]
   \caption{ RAPM (our)}
   \label{alg:IPCM}
\begin{algorithmic}[1]
   \State \textbf{Input}: $\boldsymbol{\theta}$: teacher model, $\{t_n\}_{n=N}^{0}$, $\Delta$
, $\{w_n\}_{n=N-1}^0$ \State $\;\;\;\;\;\;\;\;\;$ $\boldsymbol{\varphi}$: LoRA parameters in student model  
\State $\;\;\;\;\;\;\;\;\;$ $D_{\boldsymbol{\phi}_i}$, $i=1,2$ : discriminator 1 and 2  
   \State \textbf{for} $k=0,1,\ldots,K$ 
   \State $\;\;$ loss = 0 
    \State $\;\;$ Sample $\boldsymbol{z}_N=\tilde{\boldsymbol{z}}_N\sim \mathcal{N}(\textbf{0}; \sigma_{t_N}^2\boldsymbol{I})$
    \State $\;\;$ \textbf{for} $n=N-1$ to 0 do 
    \State $\;\;\;\;$ $\hat{\boldsymbol{z}}_{n+1}^{\Delta}\leftarrow \textrm{DDIM}(\tilde{\boldsymbol{z}}_{n+1}, t_{n+1}, t_{n+1}-\Delta, 1; \boldsymbol{\theta})$ 
    \State $\;\;\;\;$ $\hat{\boldsymbol{z}}_{n}^{\boldsymbol{\varphi}}\leftarrow \textrm{DDIM}(\hat{\boldsymbol{z}}_{n+1}^{\Delta}, t_{n+1}-\Delta, t_{n}, 1; \boldsymbol{\varphi})$
    \State $\;\;\;\;$ $\hat{\boldsymbol{z}}_{n}\leftarrow \textrm{detach}(\hat{\boldsymbol{z}}_{n}^{\boldsymbol{\varphi}})$ \hspace{20mm} \textcolor{blue}{[1st detach]}
\State $\;\;\;\;$ $\boldsymbol{z}_n^{\boldsymbol{\varphi}}\leftarrow \textrm{DDIM}(\boldsymbol{z}_{n+1}, t_{n+1}, t_n, 1; \boldsymbol{\varphi})$    
    \State $\;\;\;\;$ $\textrm{loss }+\hspace{-1mm}= H(\boldsymbol{z}_n^{\boldsymbol{\varphi}}, \hat{\boldsymbol{z}}_{n})+D_{\boldsymbol{\phi}_1}(\boldsymbol{z}_n^{\boldsymbol{\varphi}}, \hat{\boldsymbol{z}}_{n}|t_n) $
    \State $\;\;\;\;$ $\tilde{\boldsymbol{z}}_n\leftarrow \textrm{DDIM}(\tilde{\boldsymbol{z}}_{n+1}, t_{n+1}, t_n, M; \boldsymbol{\theta})$
    \State $\;\;\;\;$ $\textrm{loss } +\hspace{-1mm}= w_n\left[ H(\boldsymbol{z}_n^{\boldsymbol{\varphi}}, \tilde{\boldsymbol{z}}_n)+D_{\boldsymbol{\phi}_2}(\boldsymbol{z}_n^{\boldsymbol{\varphi}}, \tilde{\boldsymbol{z}}_n| t_n) \right]$
    \State $\;\;\;\;$ ${\boldsymbol{z}}_{n}\leftarrow \textrm{detach}(\boldsymbol{z}_n^{\varphi})$ \hspace{20mm}  \textcolor{blue}{[2nd detach]}
    \State $\;\;$  \textbf{end for}
    \State $\;\;$ \textbf{if} $\textrm{mod}(k,2)=0$
    \State $\;\;\;\;$ $\boldsymbol{\phi}_i\leftarrow \boldsymbol{\phi}_i - \eta\nabla_{\boldsymbol{\phi}_i }  \textrm{loss}$, $i=1,2$
    \State \;\;\textbf{else}
    \State $\;\;\;\;$ $\boldsymbol{\varphi}\leftarrow \boldsymbol{\varphi} - \eta\nabla_{\boldsymbol{\varphi} }  \textrm{loss}$ 
  \State $\;\;$ \textbf{end if}
  \State  \textbf{end for}
\end{algorithmic}
\vspace{-0.5mm}
\end{algorithm}

\subsection{Design of RAPM}

\textbf{Motivation}: Our primary objective is to propose a high-performance diffusion distillation method that is able to operate on a single GPU. As noted in Table~\ref{tab:GPU_bachsize}, PCM utilized 8 A800 GPUs to accommodate training data of batchsize 160. One notable property of PCM is that the method takes the clean image contaminated with Gaussian noise as input to the teacher and student diffusion models to construct loss functions. Intuitively speaking, a sufficient number of training samples are required in PCM to capture the statistics of the noisy data in the forward diffusion process.    

To reduce the batchsize, one natural strategy is for the student model to closely follow the fine-grained sampling trajectories of the teacher model with a set of coarse timesteps $\{t_n\}_{n=N}^{0}$. In doing so, the same initial Gaussian noise can be taken as input to both the student and teacher models in generation of the two diffusion trajectories. See also \cite{Zhou24SFD} for a similar motivation. 

\begin{figure}[t!]
\centering
\hspace{-0mm}\includegraphics[width=95mm]{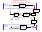} 
\vspace*{-0.2cm}
\caption{\small Demonstration of the functional losses being constructed in RAPM. Two sources of losses are introduced in RAPM, one for matching the  relative diffusion position $\hat{\boldsymbol{z}}_n$ and the other one for matching the absolute diffusion position $\tilde{\boldsymbol{z}}_n$ . } 
\label{fig:RAPM_demo}
\vspace*{-0.0cm}
\end{figure}

Let us first consider the computation of the teacher diffusion trajectories at the set of coarse timesteps $\{t_n\}_{n=N}^{0}$. To achieve a high accuracy integration approximation, we perform $M$-step DDIM within each coarse time-slot. Suppose the initial Gaussian noise is $\tilde{\boldsymbol{z}}_N\sim \mathcal{N}(\textbf{0}; \sigma_{t_N}^2\boldsymbol{I})$. We can then compute the diffusion positions $\{\tilde{\boldsymbol{z}}_n\}_{n=N-1}^{0}$ of the teacher model sequentially by applying $M$-step DDIM once per index $n$, represented as 
\begin{align}
\tilde{\boldsymbol{z}}_n = \textrm{DDIM}(\tilde{\boldsymbol{z}}_{n+1}, t_{n+1}, t_n, M; \boldsymbol{\theta}) \quad n=N-1, \ldots, 0.
\label{equ:RAPM_T}
\end{align}
See the upper part of Fig.~\ref{fig:RAPM_demo} for a demonstration of the teacher trajectory. To save training time for the student model in practice, one can pre-compute and store a certain number (e.g., 1000) of the teacher diffusion trajectories. Since both $N$ and the size of $\tilde{\boldsymbol{z}}_n$ (e.g., $4\times 128\times 128$ in SDXL backbone) are reasonably small, the above process does not consume much storage space. When the student model is to be trained, the teacher trajectories can be effectively loaded into memory.  

The next step is to first compute the student trajectory starting with the same initial Gaussian noise $\boldsymbol{z}_N=\tilde{\boldsymbol{z}}_N$, and then to construct the loss functions. As we mentioned earlier, the loss functions will be built by matching the relative and absolute diffusion positions.

\noindent\textbf{Matching the relative diffusion positions}: Without loss of generality, let us focus on the time-slot $[t_{n+1}, t_{n}]$. Inspired by the loss function in PCM, we first compute the student diffusion position $\boldsymbol{z}_n^{\boldsymbol{\varphi}}$ at timestep $t_n$ by applying 1-step DDIM with $\boldsymbol{z}_{n+1}$, given by 
\begin{align}
\boldsymbol{z}_n^{\boldsymbol{\varphi}} &= \textrm{DDIM}(\boldsymbol{z}_{n+1}, t_{n+1}, t_n, 1; \boldsymbol{\varphi}),  
\label{equ:RAPM_S1}
\end{align}
where the student model $\boldsymbol{\varphi}$ is to be learned. 

\begin{figure*}[t!]
\centering
\includegraphics[width=140mm]{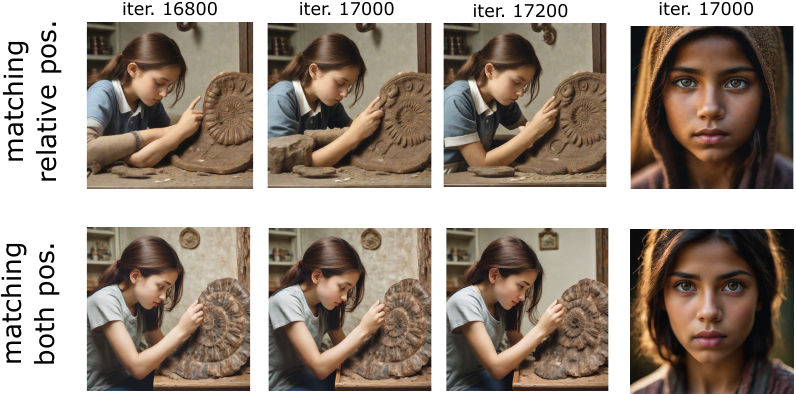} 
\vspace*{-0.0cm}
\caption{\small Qualitive comparison of RAPM with different loss functions over certain iterations when performing distillation over SDXL. The images on the top row are obtained by only matching the relative diffusion positions via the loss function (\ref{equ:RAPM_S6}). On the other hand, the images on the bottom are generated by matching both the relative and absolute positions via (\ref{equ:loss_all}).  The number of timesteps was set to 4. See Table~\ref{tab:prompts_IPCM_loss} for the associated text prompts. } 
\label{fig:RAPM_losses}
\vspace*{-0.2cm}
\end{figure*}

Similarly to PCM,  we compute a more accurate diffusion position $\hat{\boldsymbol{z}}_n$ to guide the update of the student model. In particular, $\hat{\boldsymbol{z}}_n$ is computed by applying two 1-step DDIM by the teacher and the frozen student model over $[t_{n+1}, t_{n+1}{-}\Delta]$ and $[t_{n+1}{-}\Delta, t_n]$ sequentially, given by 
\begin{align}
\hat{\boldsymbol{z}}_{n+1}^{\Delta} &= \textrm{DDIM}(\tilde{\boldsymbol{z}}_{n+1}, t_{n+1}, t_{n+1}{-}\Delta, 1; \boldsymbol{\theta}) \label{equ:RAPM_S2} \\
\hat{\boldsymbol{z}}_{n}^{\boldsymbol{\varphi}} &= \textrm{DDIM}(\hat{\boldsymbol{z}}_{n+1}^{\Delta}, t_{n+1}{-}\Delta, t_{n}, 1; \boldsymbol{\varphi}) \label{equ:RAPM_S3} \\
\hat{\boldsymbol{z}}_{n} &= \textrm{detach}(\hat{\boldsymbol{z}}_{n}^{\boldsymbol{\varphi}}). \label{equ:RAPM_S4}
\end{align}
It is worth noting that the teacher diffusion position $\tilde{\boldsymbol{z}}_{n+1}$ is used in computation of $\hat{\boldsymbol{z}}_{n}$ while $\boldsymbol{z}_n^{\boldsymbol{\varphi}} $ in (\ref{equ:RAPM_S1}) is obtained by employing $\boldsymbol{z}_{n+1}$. In principle, the teacher diffusion position  $\tilde{\boldsymbol{z}}_{n+1}$ should carry more insightful image information than the student position $\boldsymbol{z}_{n+1}$.

\begin{figure}[t!]
\centering
\hspace{-0mm}\includegraphics[width=70mm,trim=0 0.3mm 0 0,clip]{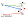} 
\caption{\small A simplified schematic diagram of our method from Fig.~\ref{fig:RAPM_demo}. The detach operation is performed to obtain $\boldsymbol{z}_{n+1}$ before computing $\boldsymbol{z}_n^{\boldsymbol{\varphi}}$ to be able to make the position matching process decoupled across different time-slots. } 
\label{fig:RAPM_demo_sim}
\vspace*{-0.0cm}
\end{figure}

Next we argue that $\hat{\boldsymbol{z}}_{n}$ computed via (\ref{equ:RAPM_S2})-(\ref{equ:RAPM_S4}) is in fact a relative diffusion position. This is because the computation of $\hat{\boldsymbol{z}}_{n}$ depends on the most recent student model $\boldsymbol{\varphi}$. Therefore, in the training process for the student model, $\hat{\boldsymbol{z}}_{n}$ keeps changing over iterations (see Fig.~\ref{fig:RAPM_demo_sim}). We will talk about the side effect for only matching the relative positions below right after introducing the cost function $\textrm{loss}^r()$.

Similarly to (\ref{equ:PCM_loss}), we let the cost function over the time-slot $[t_{n+1},t_n]$ to be
\begin{align}
\textrm{loss}_n^r(\boldsymbol{\varphi}, \boldsymbol{\phi}_1) &= H(\boldsymbol{z}_n^{\boldsymbol{\varphi}}, \hat{\boldsymbol{z}}_{n})+D_{\boldsymbol{\phi}_1}(\boldsymbol{z}_n^{\boldsymbol{\varphi}}, \hat{\boldsymbol{z}}_{n}|t_n),
\label{equ:RAPM_S5}
\end{align}
where the superscript $r$ stands for the \emph{relative position} of $\hat{\boldsymbol{z}}_{n}$.  Summing (\ref{equ:RAPM_S5}) over all index $n$ produces 
\begin{align}
\textrm{loss}^r (\boldsymbol{\varphi}, \boldsymbol{\phi}_1) = \sum_{n=0}^{N-1}\textrm{loss}_n^r(\boldsymbol{\varphi}, \boldsymbol{\phi}_1). 
\label{equ:RAPM_S6}
\end{align}
See the bottom and middle parts of Fig.~\ref{fig:RAPM_demo} for  visualization of the student trajectory and computation of $\textrm{loss}_n^r(\boldsymbol{\varphi}, \boldsymbol{\phi}_1)$.

We note that both the two diffusion positions $\boldsymbol{z}_n^{\boldsymbol{\varphi}}$ and $ \hat{\boldsymbol{z}}_{n}$ in (\ref{equ:RAPM_S5}) are dependent on the student model which is changing over training iterations. As a result, the training by only matching the relative positions might not be stable. As an example, the images in the top row of Fig.~\ref{fig:RAPM_losses} show that the image regions portraying the arms of the girl change significantly over iterations, which is undesirable. One can also observe that the images on the top row are somewhat dark. The objects in the images are not very sharp. The above issues can be largely alleviated by additionally matching the absolute diffusion positions as will be presented below.

\noindent\textbf{Matching the absolute diffusion positions}: We note that in addition to the relative diffusion position $\hat{\boldsymbol{z}}_{n}$, the absolute diffusion position $\tilde{\boldsymbol{z}}_n$ from the teacher model can also be used to guide the update of the student model. Accordingly, we propose the following cost function for measuring the difference between $\boldsymbol{z}_n^{\boldsymbol{\varphi}}$ and $ \tilde{\boldsymbol{z}}_n$ over the time-slot $[t_{n+1}, t_n]$:
\begin{align}
\textrm{loss}_n^a(\boldsymbol{\varphi}, \boldsymbol{\phi}_2; w_n) &=w_n \left[ H(\boldsymbol{z}_n^{\boldsymbol{\varphi}}, \tilde{\boldsymbol{z}}_n)+ D_{\boldsymbol{\phi}_2}(\boldsymbol{z}_n^{\boldsymbol{\varphi}}, \tilde{\boldsymbol{z}}_n| t_n)\right],
\end{align}
where $w_n\geq 0$ is a weighting factor, and the superscript $a$ stands for \emph{absolute position} of $\tilde{\boldsymbol{z}}_n$. $D_{\boldsymbol{\phi}_2}$ denotes the second discriminator. 

Finally, the overall loss is summarized as 
\begin{align}
\textrm{loss}(\boldsymbol{\varphi}, \boldsymbol{\phi}_1, \boldsymbol{\phi}_2) = \textrm{loss}_n^r(\boldsymbol{\varphi}, \boldsymbol{\phi}_1) + \sum_{n=0}^{N-1}\textrm{loss}_n^a(\boldsymbol{\varphi}, \boldsymbol{\phi}_2; w_n),
\label{equ:loss_all}
\end{align}
which is a function of the student model, and the two discriminators.

Let us consider Fig.~\ref{fig:RAPM_losses} again. The images in the bottom row are obtained by matching both the relative and absolute positions in the training process. It is clear that the images are sharper and brighter than those in the top row. Also, the image content regarding the arms of the girl in the bottom row is roughly fixed over iterations. These desirable properties can be clearly explained by the use of the absolute diffusion positions in the training process. This example indicates that it is beneficial to match both the relative and absolute diffusion positions.    

From a high level point of view, the relative diffusion position $\hat{\boldsymbol{z}}_n$ provides additional information that is missing in the absolute position $\tilde{\boldsymbol{z}}_n$ (see Fig.~\ref{fig:RAPM_demo_sim}). In principle, the additional information will be helpful to guide the update of the student model effectively. By inspection of the images in the top row of Fig.~\ref{fig:RAPM_losses}, we can conclude that use of the relative diffusion position alone already provides reasonable image quality.

\noindent\textbf{Brief discussion of the neural architectures in RAPM}: In this paper, we use the same model architectures as in \cite{Wang24PCM} for both the student model and the two discriminators per teacher model. That is, we also introduce proper LoRA adapters into a particular teacher model in designing the corresponding student model and the two discriminators. To save training memory, the two discriminators can share the same backbone in memory. 
For detailed information on the LoRA adapters,  we refer the readers to \cite{Wang24PCM}.

\subsection{Impact of the two detach operations in RAPM}
As shown in Alg.~\ref{alg:IPCM}, the update procedure of RAPM has two detach operations. The first one intends to treat the relative position $\hat{\boldsymbol{z}}_n=\textrm{detach}(\hat{\boldsymbol{z}}_n^{\boldsymbol{\varphi}})$ as a constant tensor when optimizing the student model. If the above detach operation is removed,  the relative position $\hat{\boldsymbol{z}}_n^{\boldsymbol{\varphi}}$ cannot serve as an anchor point anymore when updating the model. This is simply because $\hat{\boldsymbol{z}}_n^{\boldsymbol{\varphi}}$ will also be automatically updated as a function of the student model, which is undesirable.   

The second detach operation $\boldsymbol{z}_n=\textrm{detach}(\boldsymbol{z}_n^{\boldsymbol{\varphi}})$ in Alg.~\ref{alg:IPCM} (or equivalently $\boldsymbol{z}_{n+1}=\textrm{detach}(\boldsymbol{z}_{n+1}^{\boldsymbol{\varphi}})$ in Fig.~\ref{fig:RAPM_demo_sim}) is introduced to make the position matching be decoupled across different time-slots. From a high-level perspective, this has a similar effect as PCM, which aims to enforce consistency within each time-slot (see Alg.~\ref{alg:PCM} for details). Intuitively speaking, removal of the second detach operation might slow down the training convergence as the position matching become deeply coupled across different time-slots.

\begin{algorithm}[tb]
   \caption{ SFD \cite{Zhou24SFD} }
   \label{alg:SFD}
\begin{algorithmic}[1]
\State \textbf{Input}: $\boldsymbol{\theta}$: teacher model, $N$, $\eta$
\State $\;\;\;\;\;\;\;\;\;$ $\boldsymbol{\psi}$: student model   
   \State \textbf{repeat} 
   \State $\;\;$ Sample $\boldsymbol{z}_N=\tilde{z}_N\sim \mathcal{N}(\boldsymbol{0}, \sigma_{t_N}^2\boldsymbol{I})$ 
   \State $\;\;$ \textbf{for} $n=N-1$ \textbf{to} 0 \textbf{do}
   \State $\;\;\;\;$ $\boldsymbol{z}_{n}^{\boldsymbol{\varphi}}\leftarrow \textrm{Euler}(\boldsymbol{z}_{n+1}, t_{n+1},t_n,1;\boldsymbol{\psi})$ $\;\;$ \textcolor{blue}{[student traj.]}
   \State $\;\;\;\;$ $\tilde{\boldsymbol{z}}_{n}\leftarrow \textrm{Solver}(\tilde{\boldsymbol{z}}_{n+1}, t_{n+1},t_n,M;\boldsymbol{\theta})$ \textcolor{blue}{[teacher traj.]}
   \State $\;\;\;\;$ $\boldsymbol{\psi}\leftarrow \boldsymbol{\psi} - \eta \nabla_{\boldsymbol{\psi}} d({\boldsymbol{z}}_n^{\boldsymbol{\psi}},\tilde{\boldsymbol{z}}_n) $
   \State $\;\;\;\;$ $\boldsymbol{z}_n\leftarrow \textrm{detach}(\boldsymbol{z}_{n}^{{\boldsymbol{\psi}}})$
   \State $\;\;$ \textbf{end for}
   \State \textbf{until} convergence
\end{algorithmic}
\vspace{-0.5mm}
\end{algorithm}

\subsection{On relationship between RAPM and SFD}
The SFD method \cite{Zhou24SFD} is  recently proposed for effective diffusion distillation. Its update procedure is provided in Alg.~\ref{alg:SFD}. Similarly to our method RAPM, SFD also attempts to mimic the teacher trajectory $\{\tilde{\boldsymbol{z}}_n\}_{n=N-1}^{0}$, which is obtained by applying certain ODE solver $M$ times within each coarse time-slot $[t_{n+1}, t_n]$, given by 
\begin{align}
\tilde{\boldsymbol{z}}_{n}= \textrm{Solver}(\tilde{\boldsymbol{z}}_{n+1}, t_{n+1},t_n,M:\boldsymbol{\theta}).
\end{align}
Similarly to the update procedure of RAPM, the detach operation is also employed by SFD to enforce consistency within each time-slot separately. 

There are two main differences between RAPM and SFD. Firstly, SFD only attempts to match the absolute diffusion position $\tilde{\boldsymbol{z}}_n$ per time-slot $[t_{n+1}, t_n]$ while RAPM intends to match both the relative and absolute diffusion positions as discussed earlier. Secondly, SFD does not apply the LoRA adapters to reduce trainable parameters in the student model while RAPM does. 

\subsection{Limitations of RAPM} 
One limitation for using RAPM is that a set of teacher diffusion trajectories need to be pre-computed and stored. In our experiment for distilling SD V1.5 and SDXL, the computational time is acceptable (e.g., a few hours generating 1K trajectories for SD V1.5 using a single GPU).  However, it is not clear if the time complexity will be acceptable for more advanced diffusion models developed in the future. 


\section{Experimental Results}
We evaluated the performance of RAPM by distillating SD V1.5 and SDXL for the tasks of text-to-image generation. To accelerate the training process of RAPM, we first pre-computed and stored 1K teacher trajectories per teacher model by using 1K text prompts from the dataset CC3M \cite{Changpinyo21CC3M}. The parameter $M$ in alg.~\ref{alg:IPCM} was set to $M=25$ when generating the teacher trajectories. Each teacher trajectory stores only $5$ diffusion positions $\{\tilde{\boldsymbol{z}}_n\}_{n=4}^0$ in preparation for guiding the update of the student model later on. The dataset of COCO 2014 was used to compute the FID scores. 

In all our experiments below, RAPM was tested  with batchsize=1 on a single previous generation A6000 GPU. The maximum number of iterations was set to 20K in RAPM when distilling either SD V1.5 or SDXL. We adopted the open-source\footnote{ https://github.com/G-U-N/Phased-Consistency-Model } for PCM when evaluating the performance of RAPM. The training setups including the optimizers and hyper-parameters for the neural networks follow directly from those of PCM.      

\subsection{Performance comparison}

Table~\ref{tab:FID_sd15_compare} and  \ref{tab:tab:FID_sdxl_compare} summarize the obtained FID scores for RAPM in comparison to the state-of-the-art FIDs for distilling both SD V1.5 and SDXL. We can conclude from the tables that RAPM with 4 timesteps achieves comparable performance as the best method with 1 timestep.  See also Fig.~\ref{fig:RAPM_images} for generated images by using RAPM.

\begin{table}[t]
\caption{Image quality comparison with the SD V1.5 backbone on 30K prompts from COCO 2014. The values indicated by ${*}$ are from \cite{Wang24PCM}. }
\centering
\label{tab:FID_sd15_compare}
\begin{tabular}{|c|c|c|c|}
\hline
{\footnotesize Method} & {\footnotesize GPUs} & {\footnotesize \# Fwd Pass($\downarrow$)} & {\footnotesize FID($\downarrow$)} \\
     \hline
    {\footnotesize InstaFlow\cite{Liu23Instaflow}} 
& {\footnotesize 8 A100 }  &
{\footnotesize 1} & {\footnotesize $13.59^{*}$ } \\ \hline
{\footnotesize PCM \cite{Wang24PCM}} 
& {\footnotesize 8 A800 } &
{\footnotesize 1} & {\footnotesize $17.91^{*}$ } \\
     \hline 
{\footnotesize RAPM (\textbf{our})} & 
{\footnotesize 1 A6000 } &
 {\footnotesize 2} & {\footnotesize $14.46$ } \\ 
\hline
{\footnotesize RAPM (\textbf{our})} 
& {\footnotesize 1 A6000 } &
{\footnotesize 4} & {\footnotesize $13.91$ } \\
     \hline      
\end{tabular}
\end{table}

\begin{table}[t]
\caption{Image quality comparison with the SDXL backbone on 30K prompts from COCO 2014. The values indicated by ${*}$ are from \cite{Wang24PCM}.}
\centering
\label{tab:tab:FID_sdxl_compare}
\begin{tabular}{|c|c|c|c|}
\hline
{\footnotesize Method} & {\footnotesize GPUs} &{\footnotesize \# Fwd Pass($\downarrow$)} & {\footnotesize FID($\downarrow$)} \\
     \hline
     {\footnotesize SDXL-Lightning \cite{Lin24SDXLlighting} }
     & {\footnotesize 64 A100 }
& {\footnotesize 1} & {\footnotesize $19.73^{*}$}   \\ \hline
{\footnotesize PCM \cite{Wang24PCM} } &  {\footnotesize 8 A800 } &
 {\footnotesize 1} & {\footnotesize $21.23^{*}$} \\
     \hline 
{\footnotesize RAPM (\textbf{our}) } 
& {\footnotesize 1 A6000 } &
{\footnotesize 4} & {\footnotesize $19.21$} \\
     \hline  
\end{tabular}
\end{table}

\subsection{Ablation study}
It is of great interest to evaluate the performance of PCM with a very small batchsize. In this experiment, we computed the FID scores of PCM with bachsize of 4 and those of RAPM with batchsize of 1 over different iterations when distilling SD V1.5. The number of timesteps was set to $N=4$ in PCM and RAPM. Real images from the dataset CC3M were used when employing PCM to train the student model.    

\begin{figure}[t!]
\centering
\includegraphics[width=80mm]{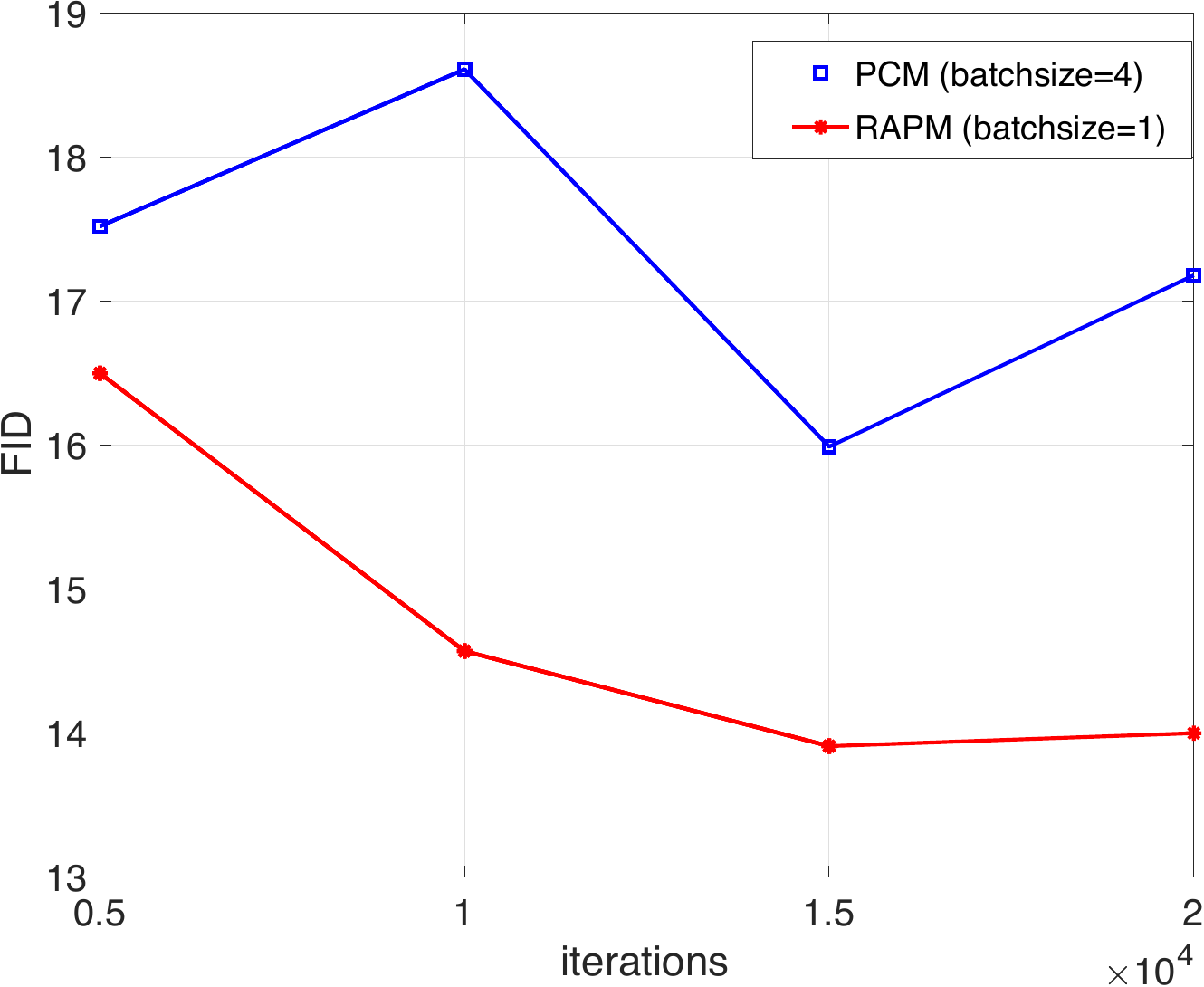}  
\vspace*{-0.0cm}
\caption{\small{ FID comparison of PCM and RAPM for distilation of SD V1.5 with 4 timesteps. } }
\label{fig:IPCM_PCM}
\vspace*{-0.2cm}
\end{figure}

Fig.~\ref{fig:IPCM_PCM} visualizes the FID curves of the two distillation methods across iterations. It is clear from the figure that RAPM performs significantly better than PCM. It is also worth noting that the FID curve of PCM fluctuates noticeably over iterations. This might be due to the fact that the setup of batchsize of 4 is too small for PCM. However, using a large batchsize for PCM may require more GPUs, which may not be accessible to some researchers. 

It is seen from Fig.~\ref{fig:IPCM_PCM} that the FID curve of RAPM is smooth across iterations even if the batchsize is set to 1. This suggests that the training process of RAPM is stable, making it a good distillation candidate in practice.

\section{Conclusions}

In this paper, we propose RAPM, a new diffusion distillation method that can train a student diffusion model with batchsize-1 on a single GPU. Instead of using real images, RAPM is designed to close the gap between the teacher and student trajectories at a set of coarse discrete timesteps, which are obtained by starting with the same initial Gaussian noises. The key step in RAPM is to align the student diffusion position with both the relative and absolute diffusion positions computed by using the teacher and student models for each time slot and iteration. While the absolute position is commonly used in other distillation methods, the relative position provides additional information to guide the update of the student diffusion model. Experimental results show that RAPM with 4 timesteps produces comparable image quality to the best method with 1 timestep using significantly smaller compute resources.



\newpage

\appendix

\onecolumn
\section{Text prompts used in the paper }

\begin{table}[h]
\caption{Text prompts used in Fig.~\ref{fig:RAPM_images} }
\centering
\label{tab:prompts_RAPM_images}
\begin{tabular}{|l|}
\hline
 ``children" \\
\hline
 $\begin{array}{l}\textrm{``a chimpanzee sitting on a wooden bench"}\end{array} $  \\
\hline   
 $\begin{array}{l}\textrm{``a goat wearing headphones"}\end{array} $  \\
\hline 
 $\begin{array}{l}\textrm{``a cat reading a newspaper"}\end{array} $  \\
\hline 
 $\begin{array}{l}\textrm{``a penguin standing on a sidewalk"}\end{array} $  \\
\hline 
 $\begin{array}{l}\textrm{``A still image of a humanoid cat posing with a hat and jacket in a bar."}\end{array} $  \\
\hline 
 $\begin{array}{l}\textrm{``a teddy bear on a skateboard in times square"}\end{array} $  \\
\hline 
 $\begin{array}{l}\textrm{``an elephant walking on the Great Wall"}\end{array} $  \\
\hline 
 $\begin{array}{l}\textrm{``A photo of an astronaut riding a horse in the forest"}\end{array} $  \\
\hline 
 $\begin{array}{l}\textrm{``a portrait of an old man"}\end{array} $  \\
\hline 
\end{tabular}
\end{table}

\begin{table}[h]
\caption{Text prompts used in Fig.~\ref{fig:RAPM_losses} }
\centering
\label{tab:prompts_IPCM_loss}
\begin{tabular}{|l|}
\hline
 ``A girl examining an ammonite fossil" \\
\hline
 $\begin{array}{l}\textrm{``portrait photo of a girl, photograph, highly detailed face, depth of field,} \\
\textrm{moody light, golden hour, style by Dan Winters, Russell James, Steve McCurry,} \\
\textrm{ centered, extremely detailed, Nikon D850, award winning photography"}\end{array}$  \\
\hline   
\end{tabular}
\end{table}

\end{document}